%% file: arxiv.tex
\documentclass[10pt,twocolumn,letterpaper]{article}

\usepackage[pagenumbers]{cvpr} 

\input{preamble}

\definecolor{cvprblue}{rgb}{0.21,0.49,0.74}
\usepackage[pagebackref,breaklinks,colorlinks,citecolor=cvprblue]{hyperref}

\def\paperID{15098} 
\def\confName{CVPR}
\def\confYear{2024}

\title{WaterHE-NeRF: Water-ray Tracing Neural Radiance Fields for Underwater Scene Reconstruction}

\author{Jingchun Zhou\textsuperscript{\rm 1},
Tianyu Liang\textsuperscript{\rm 1}, 
Dehuan Zhang\textsuperscript{\rm 1}, 
Zongxin He\textsuperscript{\rm 1}
\\
\textsuperscript{\rm 1} Dalian Maritime University
}

\begin{document}
\maketitle
\input{sec/0_abstract}    
\input{sec/1_intro}

\input{sec/2_relate}
\input{sec/3_method}

\input{sec/4_experiment}

\input{sec/5_discussion}

{
    \small
    \bibliographystyle{ieeenat_fullname}
    \bibliography{arxiv}
}


\end{document}

%% file: preamble.tex
\usepackage[dvipsnames]{xcolor}


%% file: sec/0_abstract.tex
\begin{abstract}
Neural Radiance Field (NeRF) technology demonstrates immense potential in novel viewpoint synthesis tasks, due to its physics-based volumetric rendering process, which is particularly promising in underwater scenes. Addressing the limitations of existing underwater NeRF methods in handling light attenuation caused by the water medium and the lack of real Ground Truth (GT) supervision, this study proposes WaterHE-NeRF. We develop a new water-ray tracing field by Retinex theory that precisely encodes color, density, and illuminance attenuation in three-dimensional space. WaterHE-NeRF, through its illuminance attenuation mechanism, generates both degraded and clear multi-view images and optimizes image restoration by combining reconstruction loss with Wasserstein distance. Additionally, the use of histogram equalization (HE) as pseudo-GT enhances the network’s accuracy in preserving original details and color distribution. Extensive experiments on real underwater datasets and synthetic datasets validate the effectiveness of WaterHE-NeRF. Our code will be made publicly available.


\end{abstract}

%% file: sec/1_intro.tex
\section{Introduction}
\label{sec:intro}


Neural Radiance Fields (NeRF) \cite{mildenhall2021nerf} revolutionize the representation of 3D scenes, igniting extensive research due to their ability to synthesize photorealistic perspective images. However, this initial model, tailored for aerial scenes and based on the assumption of unobstructed light transmission along rays, does not model the effects of light absorption and scattering caused by the underwater medium, limiting its applicability in the radiance field of underwater image restoration.  

Unlike aerial imaging, where light transmission is relatively unimpeded, underwater light transport experiences differential absorption across wavelengths, leading to a predominant blue-green color cast in images. Prior studies \cite{mcglamery1980computer,schechner2004clear,jaffe1990computer} have demonstrated that underwater images are principally influenced by both forward and backward scattering. Forward scattering, contributing to light absorption, induces color shifts, whereas backward scattering results in image blurriness.

\begin{figure}[t]
  \centering
   \includegraphics[width=1\linewidth]{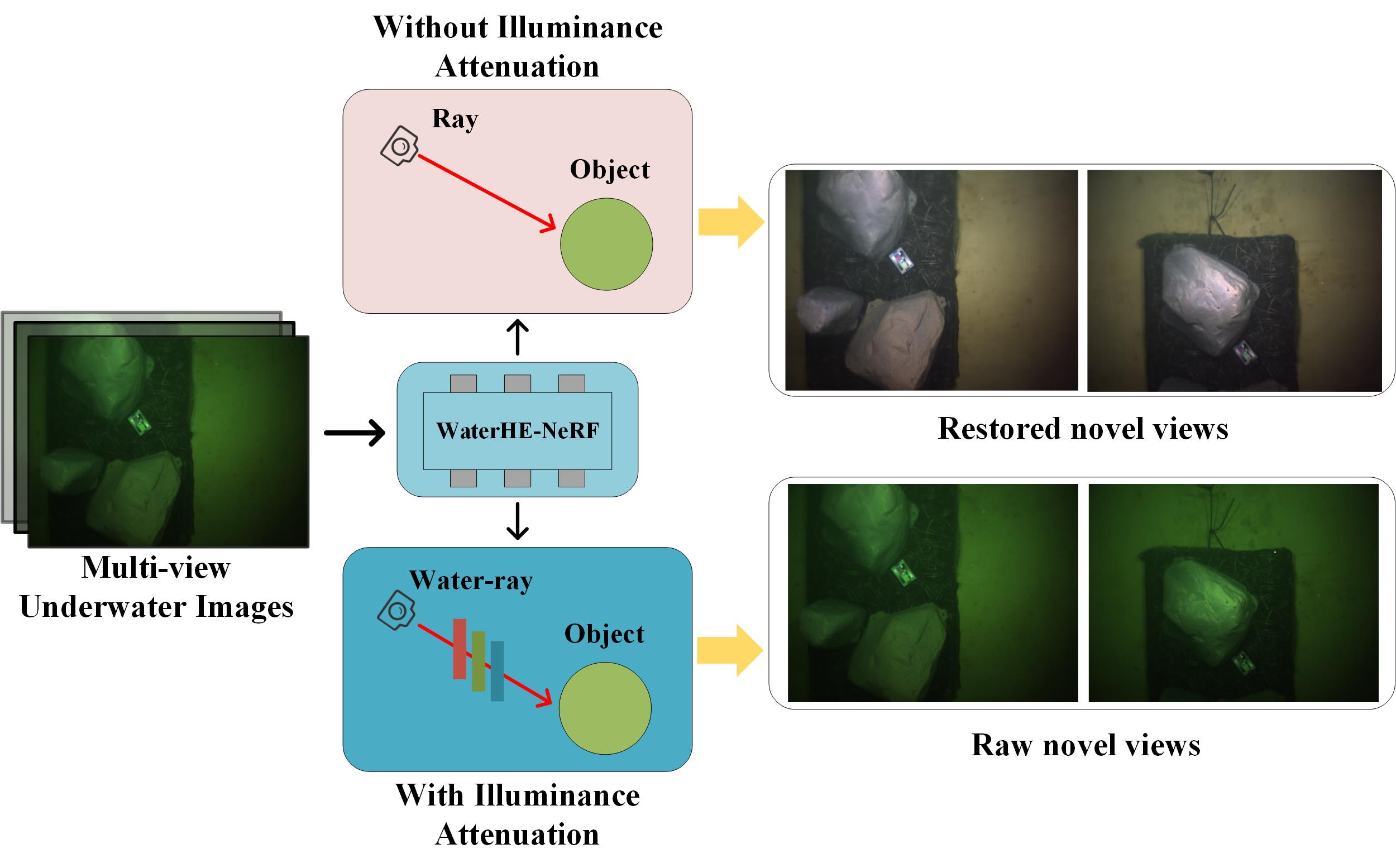}
   \caption{WaterHE-NeRF is formulated to build a radiance field with illuminance attenuation by taking multi-view underwater images as input. After training is completed, WaterHE-NeRF can synthesize novel underwater views, and synthesize their corresponding restored views by removing illuminance attenuation. }
   \label{fig:onecol}
\end{figure}

Recent some attempts \cite{levy2023seathru,sethuraman2022waternerf,zhang2023beyond} have adapted neural radiance field methods to underwater scenarios. However, these adaptations are constrained by limited recovery performance, attributed to the lack of GT images and their narrow scope of applicability. Applying 2D enhancement methods directly, as explored in \cite{drews2013transmission,akkaynak2019sea,zhou2023underwater}, overlooks the spatial information in three dimensions, thus compromising the consistency of 3D images. To address these challenges, we draw inspiration from the illuminance component in underwater physics and innovatively incorporate this aspect into NeRF's imaging process, proposing our WaterHE-NeRF model for underwater image reconstruction. This model, building upon the original NeRF \cite{mildenhall2021nerf}, which achieves excellent rendering results by estimating density and color values at each 3D point in a scene, expands the framework by integrating additional MLP networks. These networks are designed to precisely estimate illuminance attenuation at each point in a 3D scene, thereby authentically simulating underwater water-ray tracing dynamics.

During the training stage, WaterHE-NeRF employs a combination of multi-view underwater images and histogram-equalized images to construct the water-ray tracing fields. In the testing phase, WaterHE-NeRF removes the illuminance attenuation, relying on the original NeRF's radiance field for synthesizing restored images.

The key contributions of our work are as follows:




(1) We design a novel water-ray tracing field based on Retinex theory, where we firstly introduce illuminance attenuation to estimate the impact of light attenuation caused by the water medium. Our method effectively encodes color and density information in 3D space, along with illuminance attenuation. By controlling the illuminance attenuation in our rendering strategy, we are able to generate both degraded and clear multi-view images.



(2) We develop WaterHE-NeRF, which utilizes MLP networks to estimate the water-ray tracing field in scenes, synthesizing both degraded and clear images. We calculate the reconstruction loss using synthesized degraded images and real underwater images to ensure that synthetic viewpoints preserve the original image details. By solving the Wasserstein distance between the restored images and the histogram-equalization-enhanced images as the loss, we guide the restoration process to achieve the correct color distribution.

(3) We use histogram equalization (HE) enhanced images as pseudo GT. HE maintains the relative differences between pixel values during enhancement, preserving image structure and brightness. Using HE as pseudo GT helps the network robustly learn features of the original images.



%% file: sec/2_relate.tex
\section{Related Work}
\label{sec:formatting}
\subsection{Neural Radiance Fields}
NeRF model \cite{mildenhall2021nerf}, which leverages a series of images for novel-view image synthesis, has significantly influenced the field, prompting diverse research trajectories. One line of research \cite{barron2021mip,barron2022mip,hu2023tri} aims at refining the image quality produced by NeRF. Another stream \cite{chen2022tensorf,turki2023pynerf,muller2022instant,wang2023f2,fridovich2022plenoxels} seeks to accelerate the NeRF synthesis process, addressing computational efficiency. Furthermore, extending to the low-level vision domain, NeRF has proven instrumental in guiding multi-image processing tasks. This includes underwater image reconstruction \cite{levy2023seathru,sethuraman2022waternerf,zhang2023beyond}, image denoising \cite{pearl2022nan}, and low-light enhancement \cite{mildenhall2022nerf,cui2023aleth}, demonstrating its versatility and applicability across a range of imaging challenges.

Our approach is inspired by recent efforts to enhance the quality and robustness of NeRF in the low-level domain. Seathru-NeRF \cite{levy2023seathru} proposes a new model for light propagation through a scattering medium, decomposing the imaging equation into direct and backscatter components. This helps in improving the reconstruction quality of underwater images but has limitations in fully restoring them. WaterNeRF \cite{sethuraman2022waternerf} combines a physical imaging model and recovers the colors of underwater images by learning the color distribution of images processed through HE. Neural-sea \cite{zhang2023beyond} establishes a physical model for underwater robots equipped with light sources, correcting the color of underwater images by considering underwater lighting conditions and analyzing the coefficients of the underwater physical model for image restoration using images from different distances. Aleth-NeRF \cite{cui2023aleth} introduces the concept of a concealing field to simulate light attenuation in the environment, synthesizing images with normal lighting by taking low-light images as input. While these methods have made progress in specific aspects, they overlook the importance of illumination attenuation caused by water medium, leading to color biases in the restoration results. In this work, we utilize a water-ray tracing field to model the color degradation in underwater images. By learning the color distribution of histogram-equalized images, we effectively address the issue of color bias in underwater images.
\subsection{Underwater Image Enhancement}
Underwater image enhancement, a pivotal task in image processing, strives to restore images impacted by light absorption and scattering in aquatic environments to resemble those captured in air. This field has witnessed notable advancements, broadly classified into traditional and learning-based methods.

Traditional methods often leverage simplified physical models or established assumptions for image restoration. For instance, \cite{drews2013transmission} introduced the Underwater Dark Channel Prior (UDCP), mitigating the red channel's impact and accounting for water's selective light absorption. Sea-Thru \cite{akkaynak2019sea} redefines underwater light effects and applies an enhanced imaging model to underwater enhancement. \cite{zhou2023underwater} merges CIFM with prior knowledge and unsupervised techniques, enhancing monocular depth estimation accuracy and diminishing artificial illumination effects, thus achieving remarkable restoration outcomes. Despite their efficacy in certain scenarios, traditional methods often face limitations in adaptability, particularly when misaligned with specific underwater conditions, which can significantly impair restoration performance.

Conversely, learning-based approaches, empowered by extensive datasets and neural networks' robust fitting capabilities, generally demonstrate enhanced robustness and adaptability. \cite{li2020underwater} developed an end-to-end CNN-based underwater image enhancement network, trained on synthetic underwater images to restore clarity. \cite{li2017watergan} integrated GAN with a physical underwater imaging model, differentiating between synthesized and actual underwater images for image recovery. WaterNet \cite{li2019underwater} and the introduction of the Underwater Image Enhancement Benchmark Dataset (UIEBD) provide a foundation for training deep learning networks, while Ucolor \cite{li2021underwater}, inspired by underwater physical imaging, employs a medium transport-guided decoder network to enhance responses to degraded areas. Furthermore, \cite{jiang2023five} designed FA+Net, an efficient, lightweight real-time underwater enhancement network, adept at enhancing 1080p underwater images in real-time.

Despite these developments, a critical gap remains in addressing the spatial consistency of images in 3D space. Direct enhancement of reconstructed images often results in inconsistencies in image content. We bridge this gap by modifying the imaging equation of NeRF, thereby ensuring the 3D consistency of synthesized images from novel viewpoints.

%% file: sec/3_method.tex
\section{Proposed Method}

\subsection{Motivation and Problem Formulation}
The original NeRF \cite{mildenhall2021nerf} expresses the 3D scene by training a well-tuned MLP network, enabling photo-realistic novel view synthesis. By inputting the encoded 3D position $(x, y, z)$ and 2D direction $(c,\sigma )$ of points in space, the MLP network outputs the density $(\sigma)$ and pixel value $(r,g,b)$ at each point's location.

Similar to traditional image-based rendering methods, NeRF accumulates colors at points along the camera ray $r(t)=\mathbf{o} +d(t)$ to obtain the pixel value of target location $C(\mathbf{r})$. $C(\mathbf{r})$ can be represented as:
\begin{equation}\label{eq1}
C(\mathbf{r})=\int_{t_n}^{t_f}T(t)\sigma (\mathbf{r}(t) )\mathbf{c}(\mathbf{r}(t),\mathbf{d}  )dt 
\end{equation}
where integration along the ray is limited within the range of near bounds $t_n$ and far bounds $t_f$; $\sigma(\mathbf{r}(t))$ and $\mathbf{c}(\mathbf{r}(t),\mathbf{d})$ are the density and color of points along the camera ray. $T(t)$ is known as the \textit{accumulated transmittance}, which represents the probability that the ray travels from $t_n$ to $t_f$ without hitting any other particle, and is given by:
\begin{equation}\label{eq2}
T(t)=exp\left ( -\int_{t_n}^{t_f}\sigma (\mathbf{r}(s))ds   \right ) 
\end{equation}

In order to achieve differentiability in the neural network computation process, Eq. \ref{eq1}, \ref{eq2} are reformulated into their corresponding discretized form. The discrete representation of these integrals is expressed as:
\begin{equation}\label{eq3}
C(\mathbf{r} )=\sum_{i=1}^{N}T(i)(1-exp(-\sigma (\mathbf{r}(i) )\cdot \delta_i))\cdot c(\mathbf{r}(i),\mathbf{d}  ) 
\end{equation}
\begin{equation}\label{eq4}
T(i)=exp\left ( -\sum_{j=1}^{i-1}\sigma (\mathbf{r}(j) )\cdot \delta_i  \right  ) 
\end{equation}
where $\delta_i$ is the length of $i$th interval.

The differentiable NeRF model is trained by simple pixel-level reconstruction loss($L_R$):
\begin{equation}\label{eq5}
L_R=\sum_{r\in R}\left | \left | C(\mathbf{r})-\hat{C}(\mathbf{r}) \right |  \right |^{2}
\end{equation}
$L_R$ compared each rendered pixel $\hat{C}(\mathbf{r})$ to corresponding GT pixel value $C(\mathbf{r})$. Then the NeRF model can perform photo-realistic novel view synthesis.
\begin{figure}[t]
  \centering
   \includegraphics[width=1\linewidth]{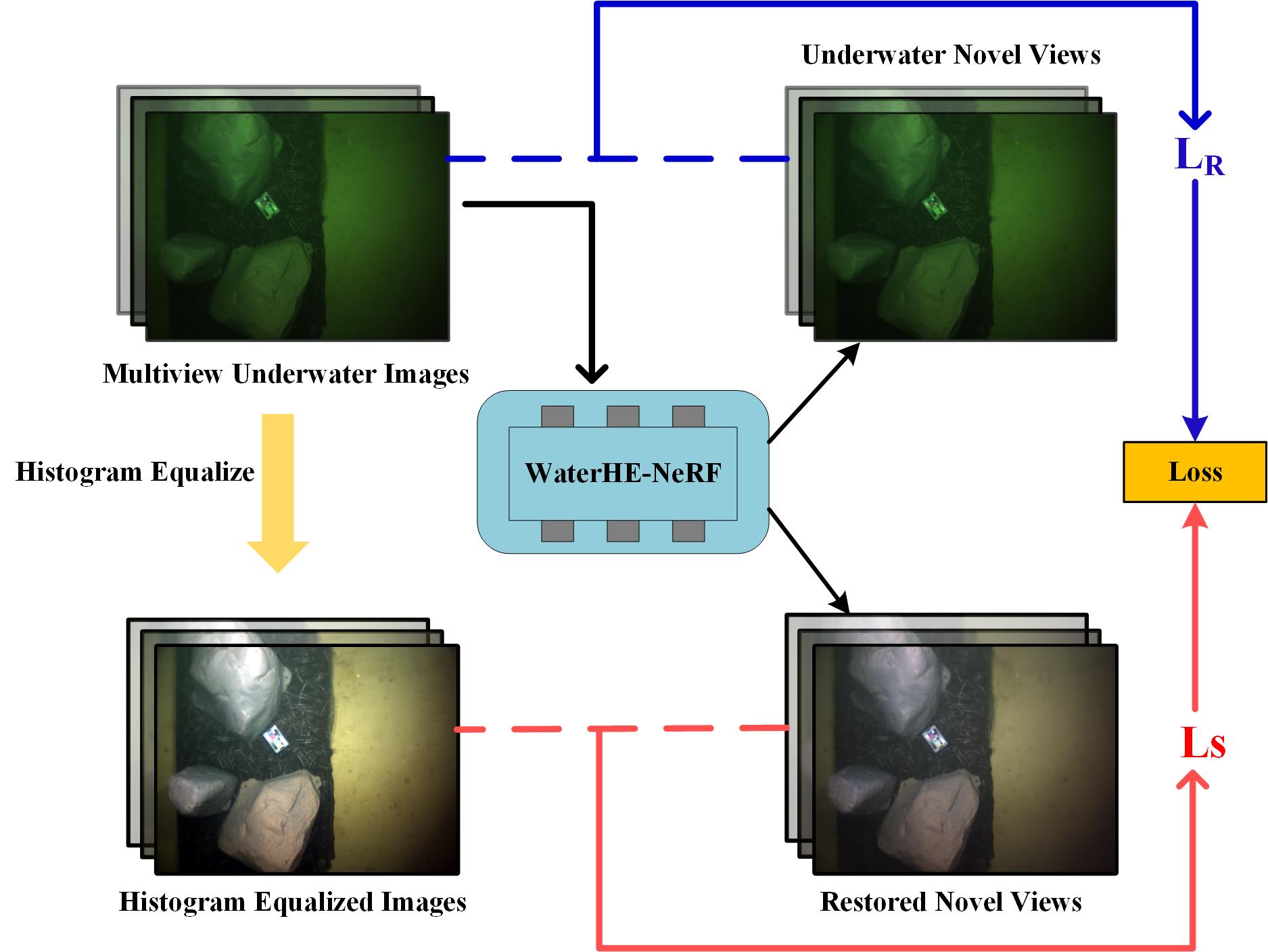}
   \caption{The overview of WaterHE-NeRF. Specifically, the model takes multi-view degraded underwater images as input, synthesizing underwater novel views and restored novel views. Then the reconstruction loss is computed between underwater views and raw input, while the color distribution loss is computed between restored views and histogram-equalized images.}
   \label{fig:twocol}
\end{figure}

However, in contrast to image formation in clear air, the absorption and scattering of light in water lead to light attenuation and deblured image details. These factors significantly undermine NeRF's capacity to precisely capture the depth and surface intricacies of underwater scenes. 


Our method is inspired by the Retinex model \cite{zhang2023rex}, underwater images are typically divided into the reflectance $R$ and illuminance $L$:
\begin{equation}\label{eq6}
I_{\lambda }(x) =  R_\lambda(x) \times L_\lambda(x)
\end{equation}
where $I_\lambda $ represents the underwater image of $\lambda $ channel, $\lambda  \in \{ R,G,B\} $. Retinex theory aims to remove the influence of the illuminance $L$ on image formation.

\begin{figure*}[t]
  \centering
   \includegraphics[width=1\linewidth]{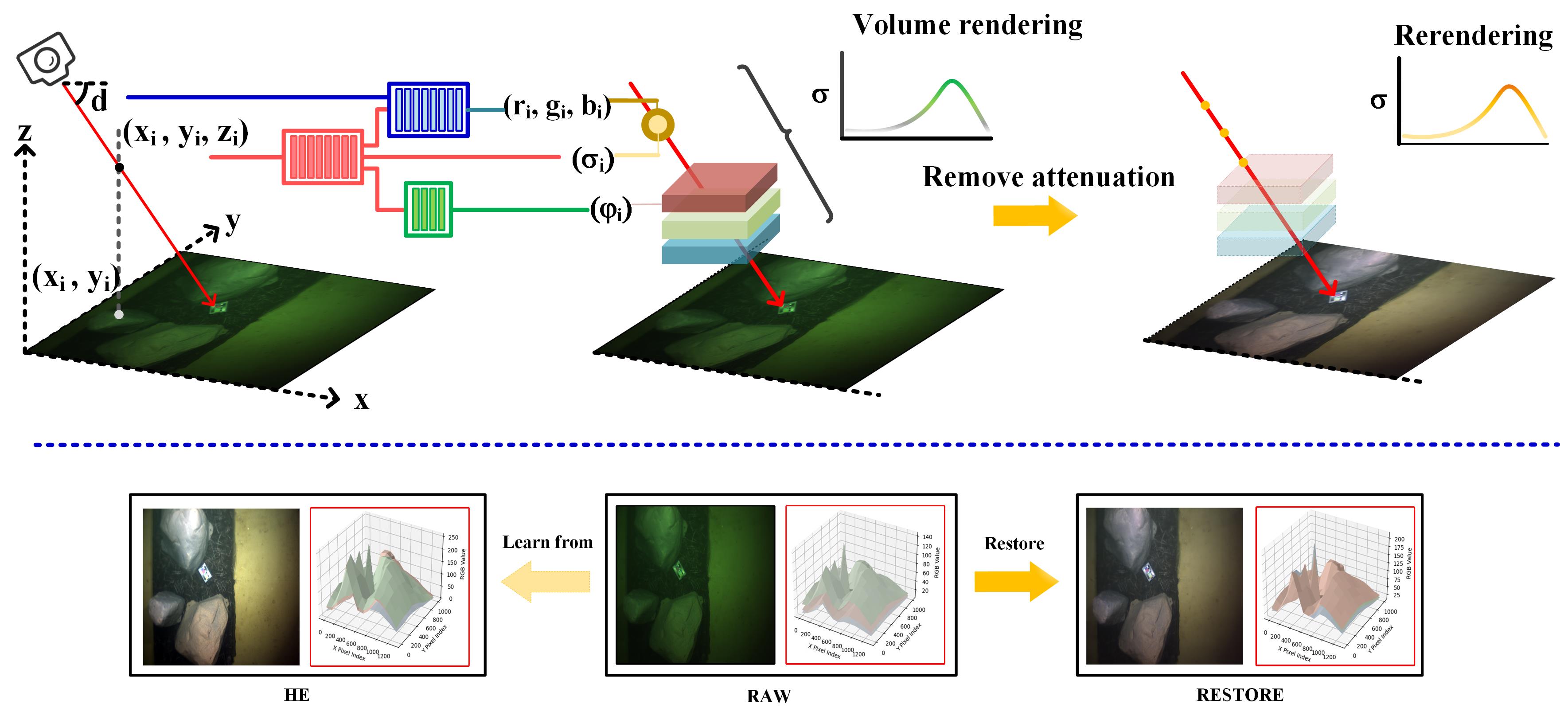}
   \caption{The pipeline of WaterHE-NeRF. The sample points on the input ray go through the MLP network to predict the pixel value, density, and illuminance attenuation of each point. In the testing stage, the illuminance attenuation is removed to rerender restored images. Our method takes the image after HE as a pseudo GT value and learns the color distribution from it to guide the restored underwater image to have the correct color distribution.}
   \label{fig:threecol}
\end{figure*}

\subsection{WaterHE-NeRF}
\label{sec:waterhe-nerf}
Rather than directly applying Eq. \ref{eq6} to the NeRF model-generated image, we made adjustments to the radiance field. This modification enables more effective handling of the influence of color-biased and blurred scenes on NeRF imaging, ensuring the model's accuracy and producing images that closely align with human visual perception. Consequently, this enhancement provides clearer visual inputs for tasks such as human and machine recognition, enhancing the overall performance of the high-level tasks.


Regarding the Eq. \ref{eq6}, we propose a novel water-ray tracing field that not only considers the color and density in the 3D space but also takes into account the illuminance attenuation $\varphi_{\lambda}(\mathbf{r}(i))$ for the underwater scene based on Retinex theory. 
$\varphi_{\lambda}(\mathbf{r}(i))$ defines the strength of a particle at location $\mathbf{r}(i)$. We add linear layers $F_{\varphi}(\cdot)$ upon the density’s part of the original NeRF’s MLP model $F_{\sigma}(\cdot)$ to obtain the illuminance attenuation value at each location $\mathbf{r}(i)$.
\begin{equation}\label{eq7}
\varphi_{\lambda}(\mathbf{r}(i))\gets F_{\varphi}\left ( F_{\sigma  }\left ( \mathbf{r}(i) \right )  \right ) 
\end{equation}

Then we use $\varphi_{\lambda}(\mathbf{r}(i))$ to calculate the attenuated accumulated transmittance $T_{\lambda}$, and is given by
\begin{equation}\label{eq8}
T_{\lambda}(i)=exp\left ( -\sum_{j=1}^{i-1}\sigma (\mathbf{r}(j) )\cdot \delta _{i}  \right )\cdot \prod_{j=1}^{i-1}\varphi_{\lambda}(\mathbf{r}(j))  
\end{equation}
while the resulting underwater pixels are obtained by:
\begin{equation}\label{eq9}
\hat{C}_{i}(r)=\sum_{i=1}^{N}T_{\lambda}(i)(1-exp(-\sigma (\mathbf{r}(i) )\cdot \delta_{i}))\cdot c(\mathbf{r}(i),\mathbf{{d}}  )  
\end{equation}
In the training stage, we use raw underwater images to calculate loss with $\hat{C}_{i}(r)$. Then, during the testing phase, we removed the illuminance attenuation and used Eq. \ref{eq3} to generate the restored image.

However, using the water-ray tracing field directly generated by the MLP to simulate the underwater medium effect can lead to an excessive influence of pseudo GT on the restoration result, thereby violating the original image's color distribution. Therefore, we average the output illuminance attenuation by batches and apply convolution to the average values to make the content of the restored image smoother.
\begin{equation}\label{eq10}
\varphi_{c}(\mathbf{r}(i) )\gets conv\left ( \frac{\sum_{i=1}^{b}\varphi_{c}(
\mathbf{r}(i) ) }{b}  \right ) 
\end{equation}
where $b$ represents the batchsize, $conv$ indicates the convolution layer.

After batch-averaging, the reconstructed images produced by the network maintain the color distribution of the raw image and are not overly influenced by pseudo GT. At the same time, the palette colors in the images also indicate that the improved method correctly restores the colors of underwater images. The results are presented in Figure \ref{fig:fourcol}.

\subsection{Water-ray Tracing Field}

The NeRF mainly focuses on the object scene information for reconstruction, represented by the reflectance $R$ in Eq. \ref{eq6}. We define each part of the two rendering Eqs. \ref{eq3}, \ref{eq9} as $\hat{C}^{R}_{i}$ and $\hat{C}^{I}_{i}$, and are given by:

\begin{equation}\label{eq11}
\hat{C}^{I}_{i}(\mathbf{r} )=T_{\lambda}^I(i)(1-exp(-\sigma (\mathbf{r}(i) )\cdot \delta _{i}))\cdot c(\mathbf{r}(i),\mathbf{{}d}  )
\end{equation}
\begin{equation}\label{eq12}
\hat{C}^{R}_{i}(\mathbf{r} )=T^R(i)(1-exp(-\sigma (\mathbf{r}(i) )\cdot \delta _{i}))\cdot c(\mathbf{r}(i),\mathbf{{}d}  )
\end{equation}
where $I$ represents the underwater image, $T_{\lambda}^I(i)$ is the transmission in $\lambda$ channel, $\lambda \in \{R,G,B\}$. Due to the selective absorption of light by the water medium, the transmission $T_{\lambda}^I(i)$ of each channel is different. $R$ indicates the clear image. Since $R$ has already compensated for the absorption in the weak channel, the transmission $T^R(i)$ is the same across all three channels.

Subsequently, we established the interrelationship between $\hat{C}^{I}_{i}$ and $\hat{C}^{R}_{i}$:
\begin{equation}\label{eq13}
\hat{C}^{I}_{i}(\mathbf{r} ) =  \hat{C}^{R}_{i}(\mathbf{r} )\cdot \prod_{j=1}^{i-1}\varphi_{\lambda}(\mathbf{r}(j) ) 
\end{equation}

For the sake of simplicity, we assume uniform attenuation strength $\varphi_{\lambda} (\mathbf{r})$ along each interval of every camera ray. Hence, Eq. \ref{eq13} is reformulated as:
\begin{equation}\label{eq14}
\hat{C}^{I}_{i}(\mathbf{r} )\approx \hat{C}^{R}_{i}(\mathbf{r} ) \cdot (\varphi_{\lambda}(\mathbf{r} ))^{i} 
\end{equation}

The Eq. \ref{eq14} is similar to the Retinex theory, aiming to eliminate the influence of the illumination component to restore the color cast of underwater images and enhance detailed information in the scene. In multi-view scenarios, synthesizing information from multiple perspectives ensures the consistency of the illumination component, robustly enhancing underwater images and preventing enhancement failure caused by scene variations.



\subsection{Loss Function}
Our loss is divided into two parts. In the training stage, we calculate the image reconstruction loss $L_{R}$ between $\hat{C}_{i}^{I}$ and underwater raw input $C(\mathbf{r})$. Furthermore, In terms of supervised images, we adopted the approach proposed in WaterNeRF. We used the Wasserstein distance \cite{arjovsky2017wasserstein} calculated with the Sinkhorn method \cite{knight2008sinkhorn} as the loss. By minimizing the Wasserstein distance between the restored image and the supervised image, we aimed to make the distribution of the restored image more closely resemble the distribution of the histogram-equalized image.
\begin{equation}\label{eq16}
L_{S}=Sinkhorn(\hat{C}_{i}^{R},C_{i}^{HE})
\end{equation}

Thus, the final loss function $L$ is a combination of $L_R$ and $L_S$, which is given by:
\begin{equation}\label{eq17}
L=L_{R}+\alpha\cdot L_{S}
\end{equation}
where $\alpha$ is the weight of $L_{S}$. We recommend setting $\alpha$ to be within the range [5e-4, 5e-5], with the specific value determined by the disparity between the results of HE and the original image.

%% file: sec/4_experiment.tex
\section{Experiments}

\begin{figure*}[tbp]
  \centering
   \includegraphics[width=1\linewidth]{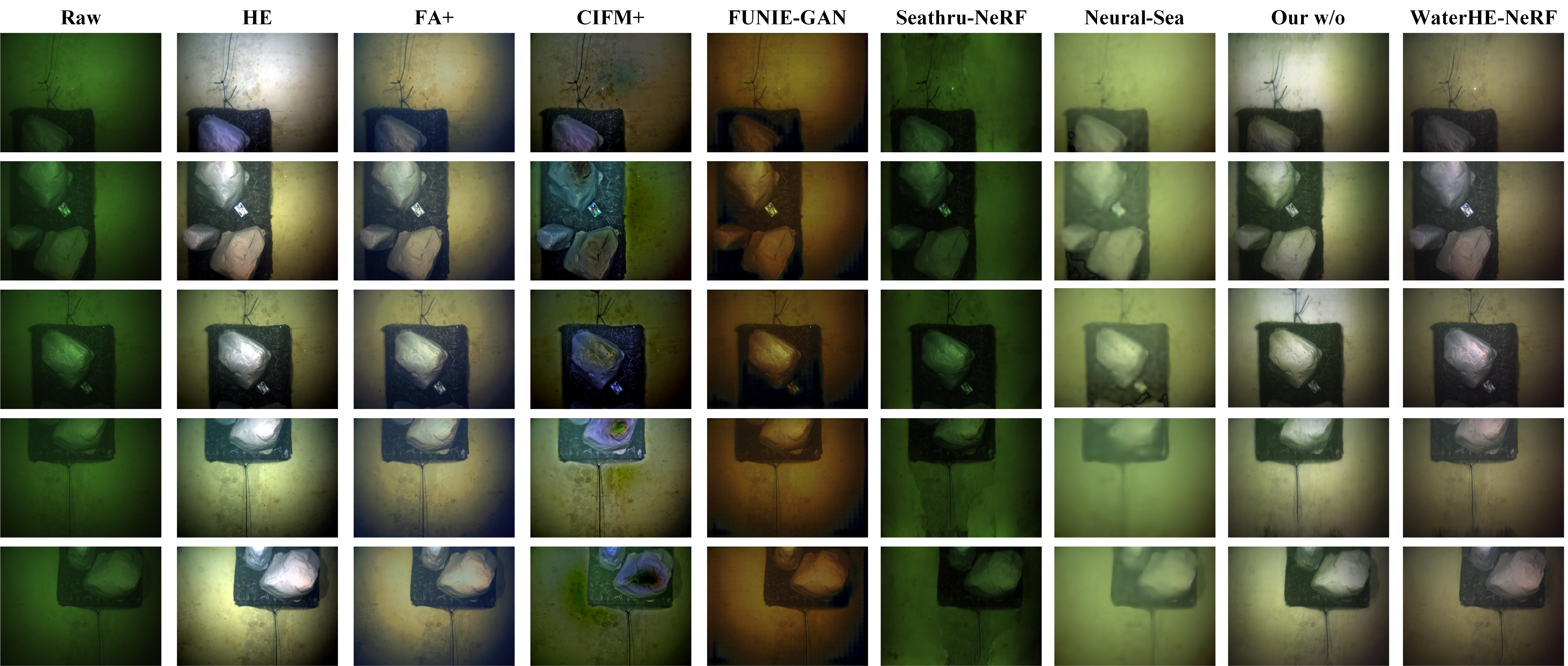}
   \caption{Scene rendering and enhanced images on the UWbundle dataset. Left to right: Raw input, Histogram Equalization \cite{pizer1987adaptive}, FA+ \cite{jiang2023five}, CIFM+ \cite{zhou2023underwater}, FUNIE-GAN \cite{islam2020fast}, Seathru-NeRF \cite{levy2023seathru}, Neural-Sea \cite{zhang2023beyond}, our method without batch-averaging, our WaterHE-NeRF result. Top to bottom: novel views and corresponding enhanced images.  }
   \label{fig:fivecol}
\end{figure*}

\subsection{Implementation Details}
We employ the ott-jax \cite{cuturi2022optimal} package, built on JAX, for the computation of the Sinkhorn loss, an essential component in optimal transport. The training leverages the Adam optimizer, with the learning rate dynamically adjusted between 5e-4 and 5e-6 across iterations. In our ray sampling process, we select 4096 camera rays per batch and uniformly sample 128 points along each ray, from near to far. WaterHE-NeRF was trained over 250,000 iterations on an Nvidia A800 tensor core GPU, equipped with 80GB of memory. The training was conducted on two distinct scenes using WaterHE-NeRF to validate its capability in synthesizing new views. For parameter tuning, $\alpha$ was set to 5e-5 for experiments on the UWbundle dataset and 5e-4 for the synthetic dataset, tailoring the model's response to each dataset's unique characteristics.

\subsection{Datasets}
In our experimental evaluation, WaterHE-NeRF was rigorously validated using two distinct datasets: the real-world UWBundle dataset and the synthetically generated LLFF-Water scene. Specifically, the LLFF-Water dataset is a synthetic underwater dataset derived from the LLFF framework\cite{mildenhall2019local}. This dual-dataset approach allowed us to comprehensively assess the performance of WaterHE-NeRF in both authentic underwater environments and controlled synthetic conditions.

In our analysis of real-world scenarios, we utilized the UWBundle dataset \cite{skinner2017automatic}, an open-source collection of images. This dataset was compiled using a controlled setup in the University of Michigan’s Marine Hydrodynamics Laboratory, where a man-made rock platform was submerged in a pure water test tank. The controlled yet realistic setting of this dataset provides an ideal environment for validating the efficacy of WaterHE-NeRF in real-world underwater conditions.

For the synthesis of the underwater scene, we employed the MipNeRF 360 model \cite{barron2022mip} to estimate depth for the well-known LLFF dataset \cite{mildenhall2019local}. The synthetic LLFF-Water dataset was then generated using Equation \ref{eq_Jaffe}, simulating underwater conditions. In this synthesis process, we meticulously set the water parameters as follows: absorption coefficients $\beta^{D}=[0.22,0.1,0.15]$, backscatter coefficients $\beta^{B}=[0.22,0.1,0.15]$, and ambient light parameters $B^{\infty}=[0.013,0.04,0.01]$. These parameter values were chosen to accurately represent typical underwater light propagation characteristics, ensuring that the synthesized LLFF-Water dataset closely mimics real underwater environments.

\begin{equation}\label{eq_Jaffe}
I_{\lambda }(x)=J_{\lambda }(x)e^{-\beta_{\lambda }^D d(x)}+B_{\lambda }\left ( 1-e^{-\beta_{\lambda }^B d(x)} \right )  
\end{equation}

\begin{figure*}[htbp]
  \centering
   \includegraphics[width=1\linewidth]{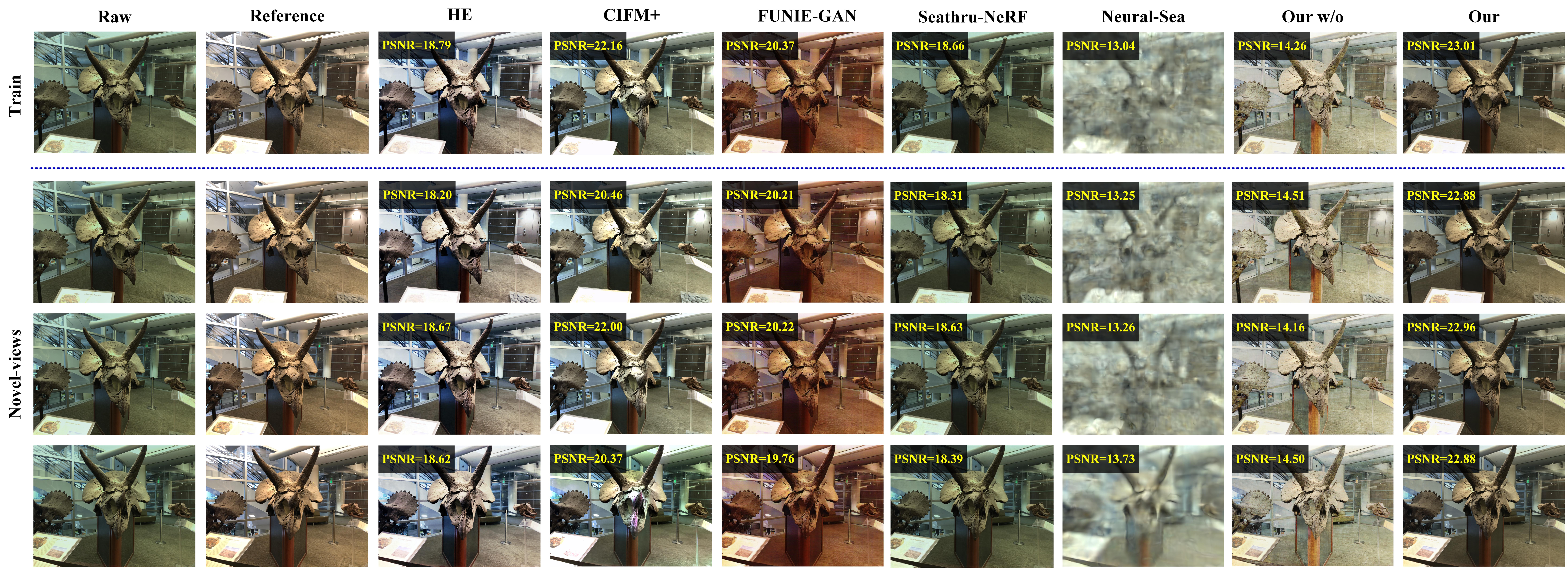}
   \caption{Experiment on synthetic Horns scene of the LLFF-water dataset. From top to bottom, we present images from different perspectives in both the training stage and the testing stage using unsupervised methods, including HE \cite{pizer1987adaptive}, CIFM+ \cite{zhou2023underwater}, FUNIE-GAN \cite{islam2020fast}, Seathru-NeRF \cite{levy2023seathru}, Neural-Sea \cite{zhang2023beyond}, our methods without batch-averaging, and our completed method.}
   \label{fig:sixcol}
\end{figure*}

\subsection{Compared Methods}
In our evaluation on the UWbundle dataset's novel views, we conducted a comprehensive comparison of our method against several SOTA methods. These include HE 
(CVGIP'1987) \cite{pizer1987adaptive}, CIFM+ (IJCV'23) \cite{zhou2023underwater}-a leading unsupervised single-image restoration technique, FUNIE-GAN (RA-L'20) \cite{islam2020fast}-an advanced unsupervised GAN-based method, FA+ (BMVC'23) \cite{jiang2023five}-a supervised deep learning network specifically designed for underwater image enhancement, and two NeRF-based models tailored for underwater scenarios: Neural-sea (RA-L'23) \cite{zhang2023beyond} and Seathru-NeRF (CVPR'23) \cite{levy2023seathru}.

To further assess the robustness and efficacy of our approach, we compared it against these methods using several reference metrics on novel views of the synthetic LLFF-Water dataset. The metrics employed for this comparative analysis included Peak Signal-to-Noise Ratio (PSNR) \cite{R23PSNR}, Structural Similarity Index Measure (SSIM) \cite{wang2004image}, Learned Perceptual Image Patch Similarity (LPIPS) \cite{zhang2018unreasonable}, and Normalized Root Mean Square Error (NRMSE). This range of metrics was chosen to provide a comprehensive assessment of image quality, structural fidelity, perceptual similarity, and error magnitude.

\subsection{Multi-view Image Enhancement Results}

\textbf{Real World Scene on UWbundle.} 
Our evaluation on UWbundle, as depicted in Fig \ref{fig:fivecol}, revealed distinctive outcomes for each method. HE, while effective in certain areas, often lead to localized over-brightness and detail loss. A notable instance is the tip of the rock appearing excessively bright and tending towards white. The CIFM+ method\cite{zhou2023underwater}, an unsupervised single-image restoration technique, faced challenges with localized abnormal color biases due to inaccuracies in depth estimation. Specifically, CIFM+\cite{zhou2023underwater} incorrectly estimated the depth at the rock's tip, resulting in inadequate color restoration and a pronounced deep green tint in that area. On the contrary, images restored using the FA+\cite{jiang2023five} demonstrated commendable overall performance. However, the method's single-image enhancement approach induced local artifacts, primarily attributable to its disregard for spatial consistency in 3D space. Consequently, the restored images by the aforementioned single-image enhancement methods exhibited a lack of 3D consistency, with noticeable content variations when observed from different perspectives.

In our unsupervised setting, Seathru-NeRF\cite{levy2023seathru} encountered difficulties in effectively correcting the pervasive color cast throughout the images. This limitation stems from its design, which does not leverage supervised image information for color correction. Conversely, Neural-Sea\cite{zhang2023beyond} tailored for mobile robots with integrated light sources, necessitates precise lighting parameters and datasets capturing variable distances to objects. As a result, its performance is suboptimal when applied to standard LLFF datasets, which do not meet these specific requirements. Our method is designed to simulate light absorption effects in water comprehensively. It effectively learns from the color distribution of images enhanced through HE, leading to more accurate restoration of colors while retaining the original image details. Notably, in terms of color accuracy, especially on color board content, our approach demonstrates superior performance, even surpassing that of supervised methods.

\noindent \textbf{Synthetic Scene on LLFF-Water.} Our comparative analysis of the unsupervised SOTA and WaterHE-NeRF on the LLFF-Water dataset is illustrated in Fig \ref{fig:sixcol} and quantitatively summarized in Table \ref{table1}. In scenarios where images are heavily affected by the color cast due to light absorption in water, Seathru-NeRF\cite{levy2023seathru} encounters challenges in effectively differentiating between the water medium and the objects. 
This limitation is reflected in its suboptimal reconstruction results. Similarly, the synthetic dataset's nature does not align with Neural-Sea's specific requirements, designed predominantly for datasets captured with precise lighting conditions and varying object distances. Consequently, Neural-Sea also exhibits less effective reconstruction on the synthetic LLFF-Water dataset. In contrast, our WaterHE-NeRF method demonstrates a proficient understanding of the color distribution in histogram-equalized images. This capability enables it to restore images closely resembling the original scenes, without being overly influenced by the pseudo GT. This balance ensures the preservation of authentic color and detail, underscoring the effectiveness of our approach in handling synthetic underwater imaging conditions.

\begin{table}
  \begin{center}
\resizebox{\linewidth}{!}{
    \begin{tabular}{c|cccc} 
      \hline
      \textbf{}& \textbf{PSNR} $\uparrow$ & \textbf{SSIM} $\uparrow$ & \textbf{LPIPS} $\downarrow$ & \textbf{NRMSE} $\downarrow$\\
      \hline
      HE (CVGIP'87)& 18.363 & 0.775  & 0.193 & 0.241 \\
      CIFM+ (IJCV'23)  & 19.288 & 0.797  & 0.180 & 0.228 \\
      FUNIE-GAN (RA-L'20) & 19.913 & 0.788  & 0.211 & 0.201 \\
      Seathru-NeRF (CVPR'23) & 18.505 & 0.818  & 0.131 & 0.236 \\
      Neural-sea (RA-L'23) & 13.594 & 0.424 & 0.794 & 0.415 \\
      WaterHE-NeRF & \textcolor{red}{22.578} & \textcolor{red}{0.841} & \textcolor{red}{0.112} & \textcolor{red}{0.148} \\
      \hline
    \end{tabular}}
\caption{PSNR, SSIM, LPIPS, and NRMSE of the unsupervised methods and WaterHE-NeRF on the LLFF-Water dataset.}
\label{table1}
  \end{center}
\end{table}

\begin{table}
  \begin{center}
\resizebox{\linewidth}{!}{
    \begin{tabular}{c|cccc} 
      \hline
      \textbf{}& \textbf{PSNR} $\uparrow$ & \textbf{SSIM} $\uparrow$ & \textbf{LPIPS} $\downarrow$ & \textbf{NRMSE} $\downarrow$\\
      \hline
      w/o & 14.546 & 0.641  & 0.253 & 0.372 \\
      WaterHE-NeRF & \textbf{22.578} & \textbf{0.841} & \textbf{0.112} & \textbf{0.148} \\
      \hline
    \end{tabular}}
\caption{The metrics of the ablation study. w/o represents the model without batch-averaging Bold indicates the best.}
\label{table_ablation}
  \end{center}
\end{table}

\begin{figure}[htbp]
  \centering
   \includegraphics[width=0.8\linewidth]{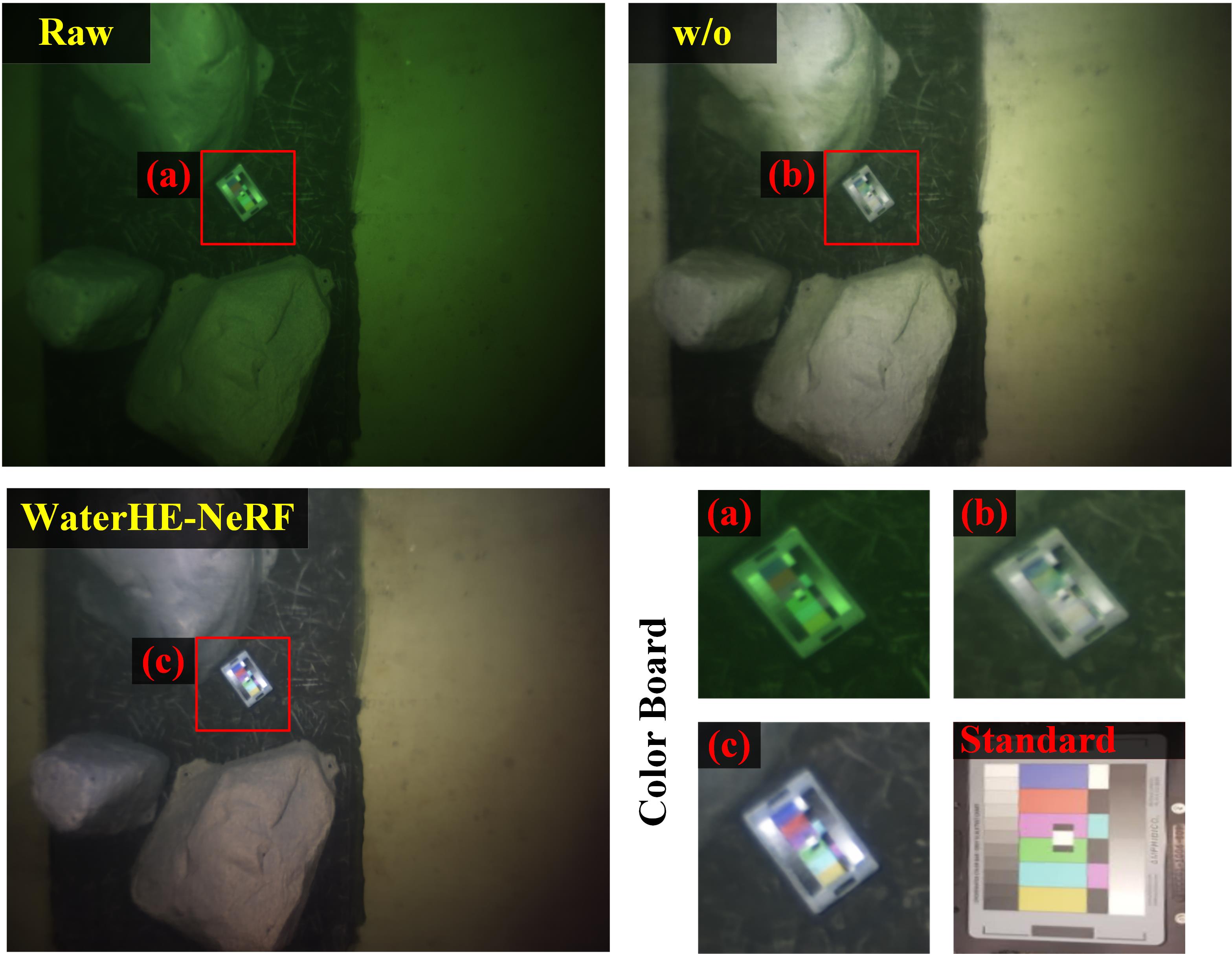}
   \caption{Ablation analysis on the UWbundle dataset. \textbf{w/o} represents the results of WaterHE-NeRF without batch-averaging, \textbf{WaterHE-NeRF} represents the results of entire WaterHE-NeRF. \textbf{(a)}, \textbf{(b)} and \textbf{(c)} show the color board results from the \textbf{Raw}, \textbf{w/o}, and \textbf{WaterHE-NeRF} respectively. \textbf{Standard} shows the color board in the air. }
   \label{fig:fourcol}
\end{figure}

\noindent \textbf{Ablation Study.}
In Sec. \ref{sec:waterhe-nerf}, we recognized that batch-averaging operations play a crucial role in enhancing the novel view recovery performance. To empirically validate this, we conducted a series of ablation experiments focusing on the impact of batch-averaging within our WaterHE-NeRF framework. The quantitative results of these experiments are detailed in Table \ref{table_ablation}, and the corresponding visual comparisons are provided in Figure \ref{fig:fourcol}. Our findings indicate that WaterHE-NeRF, with batch-averaging integrated, achieves the highest performance metrics and delivers visually superior results. Notably, when comparing against configurations without batch-averaging (denoted as w/o in the results), the color accuracy improvements are evident. This is particularly highlighted in the color board segment of Figure \ref{fig:fourcol}, where WaterHE-NeRF demonstrates a more precise recovery of color information in the scene. These results underscore the effectiveness of batch-averaging operations in our model, enhancing both the quantitative and qualitative aspects of underwater image reconstruction.

\subsection{Multi-view Consistency Results}
\begin{figure}[t]
  \centering
   \includegraphics[width=0.95\linewidth]{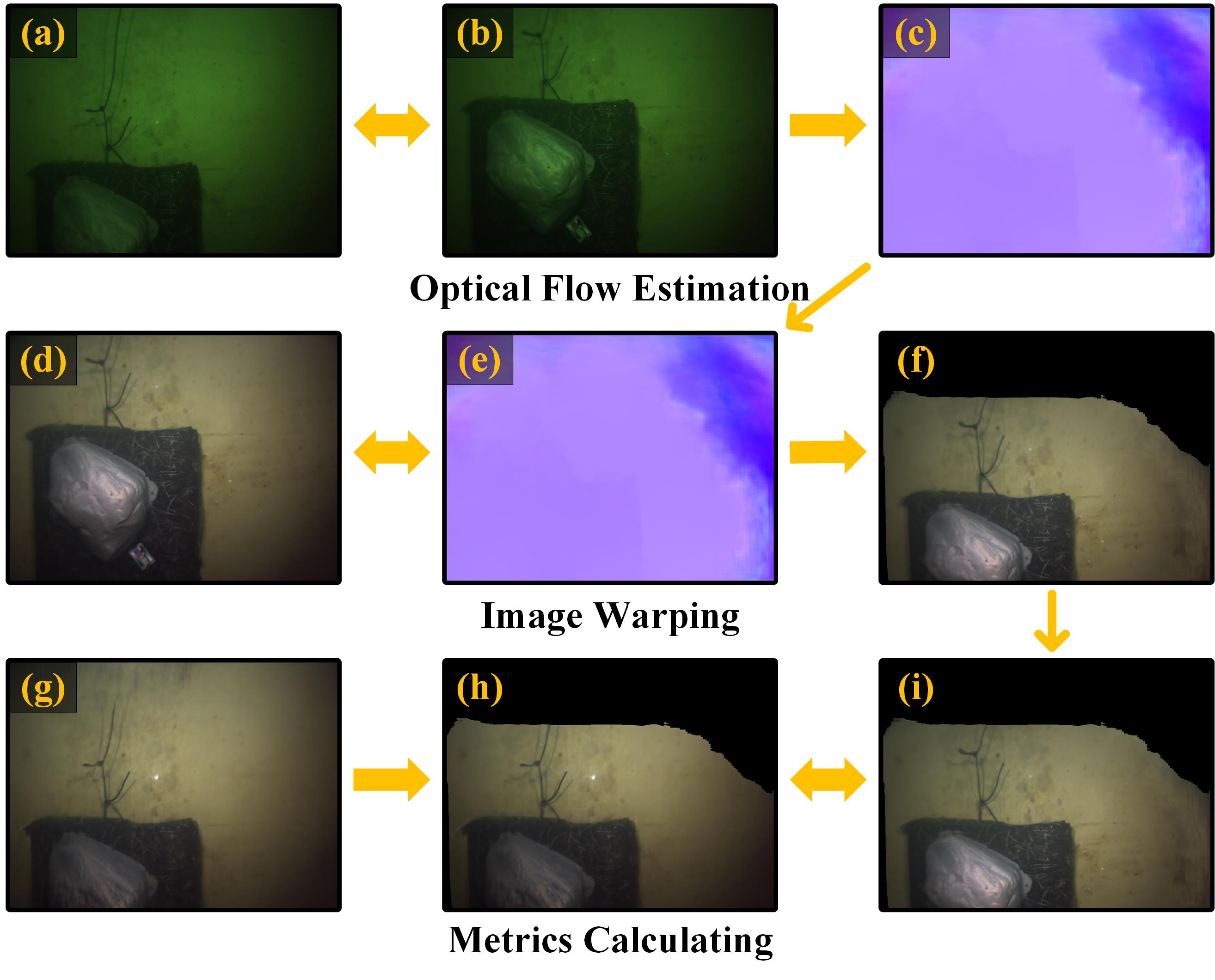}
   \caption{Workflow demonstration for measuring multi-view consistency.}
   \label{fig:sevencol}
\end{figure}

We evaluated and compared the consistency of image enhancement effects across various methods using the UWbundle dataset. The procedure for measuring multi-view consistency is depicted in Figure \ref{fig:sevencol}. Initially, we employed Raft \cite{teed2020raft} for optical flow estimation on raw images from different viewpoints, as exemplified by images (a) and (b) of the dataset. This was followed by utilizing the obtained optical flow map from (e) to perform a warp operation on the enhanced image (d), aligning it with image (g). The resulting aligned images are illustrated as (f).

\begin{table}
  \begin{center}
\resizebox{\linewidth}{!}{
    \begin{tabular}{c|cccc} 
      \hline
      \textbf{}& \textbf{PSNR} $\uparrow$ & \textbf{SSIM} $\uparrow$ & \textbf{LPIPS} $\downarrow$ & \textbf{NRMSE} $\downarrow$\\
      \hline
      HE (CVGIP'87) & 15.691 & 0.784  & 0.161 & 0.385 \\
      CIFM+ (IJCV'23) & 22.291 & 0.811  & 0.140 & 0.475 \\
      FUNIE-GAN (RA-L'20) & 23.103 & 0.817  & 0.145 & 0.396 \\
      FA+ (BMVC'23) & 19.203 & 0.856  & 0.115 & 0.352 \\
      Seathru-NeRF (CVPR'23) & \textbf{32.307} & 0.953  & 0.0553 & \textbf{0.184} \\
      Neural-sea (RA-L'23) & 21.369 & \textbf{0.959} & \textbf{0.0356} & 0.237 \\
      WaterHE-NeRF & 21.502 & 0.913 & 0.0821 & 0.323 \\
      \hline
    \end{tabular}}
\caption{PSNR, SSIM, LPIPS, and NRMSE of compared methods and WaterHE-NeRF on the aligned images of UWbundle.}
\label{table2}
  \end{center}
\end{table}

For a comprehensive analysis, we applied the same masking technique used in the warp operation to the enhanced image (g), aligning it with image (f). The effectiveness of the consistency was then assessed by comparing the processed image (h) in against the aligned image (i). This comparative analysis enabled us to quantify the consistency across multiple viewing angles, a crucial aspect of underwater image enhancement.

The summarized results, as presented in Table \ref{table2}, reveal interesting insights. Although Seathru-NeRF and Neural-sea showed promising results in SSIM, LPIPS, and NRMSE, a closer inspection, particularly in Figure \ref{fig:fivecol}, indicates that their high performance metrics are somewhat misleading, stemming from recovery failures. Specifically, images that are excessively dark or blurry tend to lose pixel details and overall image structure, resulting in anomalously high scores in these metrics.

Conversely, our WaterHE-NeRF approach emerged as the most effective, achieving the best results in terms of SSIM, LPIPS, and NRMSE. Our method not only ensures multi-view consistency but also maintains high-quality image restoration, highlighting its effectiveness in underwater image enhancement across varied perspectives.

%% file: sec/5_discussion.tex

\section{Conclusions}


In this work, we present WaterHE-NeRF, a novel framework for the consistent restoration of underwater scenes. By integrating illuminance attenuation to estimate the impact of light attenuation in water, we construct a novel water-ray tracing field based on Retinex theory. WaterHE-NeRF synthesizes both degraded and clear images using this field, calculating reconstruction loss with real images and Wasserstein loss with histogram equalization results. Comprehensive loss enables WaterHE-NeRF to achieve commendable image restoration results, using only the corresponding HE images while preserving details from the original images and maintaining multi-view consistency. However, our method has a limitation. In cases where the underwater image is excessively blurred due to backscattering, WaterHE-NeRF does not restore it well due to the limited enhancing effect of HE.
